\documentclass[letterpaper, 10 pt, conference]{ieeeconf}  

\IEEEoverridecommandlockouts                              

\usepackage{graphicx}
\usepackage{amsmath} 
\usepackage{amssymb}  
\usepackage{subcaption}
\usepackage{dblfloatfix}
\usepackage{cite}
\usepackage{booktabs}

\makeatletter
\let\NAT@parse\undefined
\makeatother
\usepackage[backref, colorlinks=true]{hyperref}
\hypersetup{
    linkcolor=blue,
    citecolor=red,
    pdftitle={2D Floor Plan Segmentation Based on Down-sampling},
    pdfauthor={Mohammadreza Sharif},
    }

\title{\LARGE \bf
2D Floor Plan Segmentation Based on Down-sampling*
}

\author{Mohammadreza Sharif, Kiran Mohan, and Sarath Suvarna$^{1}$
\thanks{*This work was supported by Neato Robotics, San Jose, CA.}
\thanks{$^{1}$Mohammadreza Sharif, Kiran Mohan, and Sarath Suvarna, are with Neato Robotics, 50 Rio Robles, San Jose, CA
        {\tt\small \{mohammadreza.sharif, kiran.mohan, sarath.suvarna\}@neatorobotics.com}}%
}

\begin{document}

\maketitle
\thispagestyle{empty}
\pagestyle{empty}

\begin{abstract}

In recent years, floor plan segmentation has gained significant attention due to its wide range of applications in floor plan reconstruction and robotics. In this paper, we propose a novel 2D floor plan segmentation technique based on a down-sampling approach. Our method employs continuous down-sampling on a floor plan to maintain its structural information while reducing its complexity. We demonstrate the effectiveness of our approach by presenting results obtained from both cluttered floor plans generated by a vacuum cleaning robot in unknown environments and a benchmark of floor plans. Our technique considerably reduces the computational and implementation complexity of floor plan segmentation, making it more suitable for real-world applications. Additionally, we discuss the appropriate metric for evaluating segmentation results. Overall, our approach yields promising results for 2D floor plan segmentation in cluttered environments.

\end{abstract}

\section{INTRODUCTION}

Floor plan segmentation has numerous applications in various fields, such as mobile robot navigation, semantic localization, virtual and augmented reality, architectural layout analysis, and real estate. The challenges associated with this problem depend on the input type, which varies for each application. The input data may include a stream of RGB-D frames, multiple panoramic 3D scans, or 2D laser range data. 2D floor plan segmentation involves segmenting 2D maps generated using a 2D laser scanner or from the projection of higher-dimensional inputs \cite{Turner2014-ev,Ambrus2017-zj}. Despite the various proposed solutions for 2D floor plan segmentation, a robust and fast algorithm is yet to be introduced. In this paper, we present a simple, robust, and fast algorithm for 2D floor plan segmentation, which can be used as a stand-alone solution or as an initialization for other segmentation problems.

\begin{figure}[!thp]
	\centering
        \includegraphics[width=0.40\textwidth]{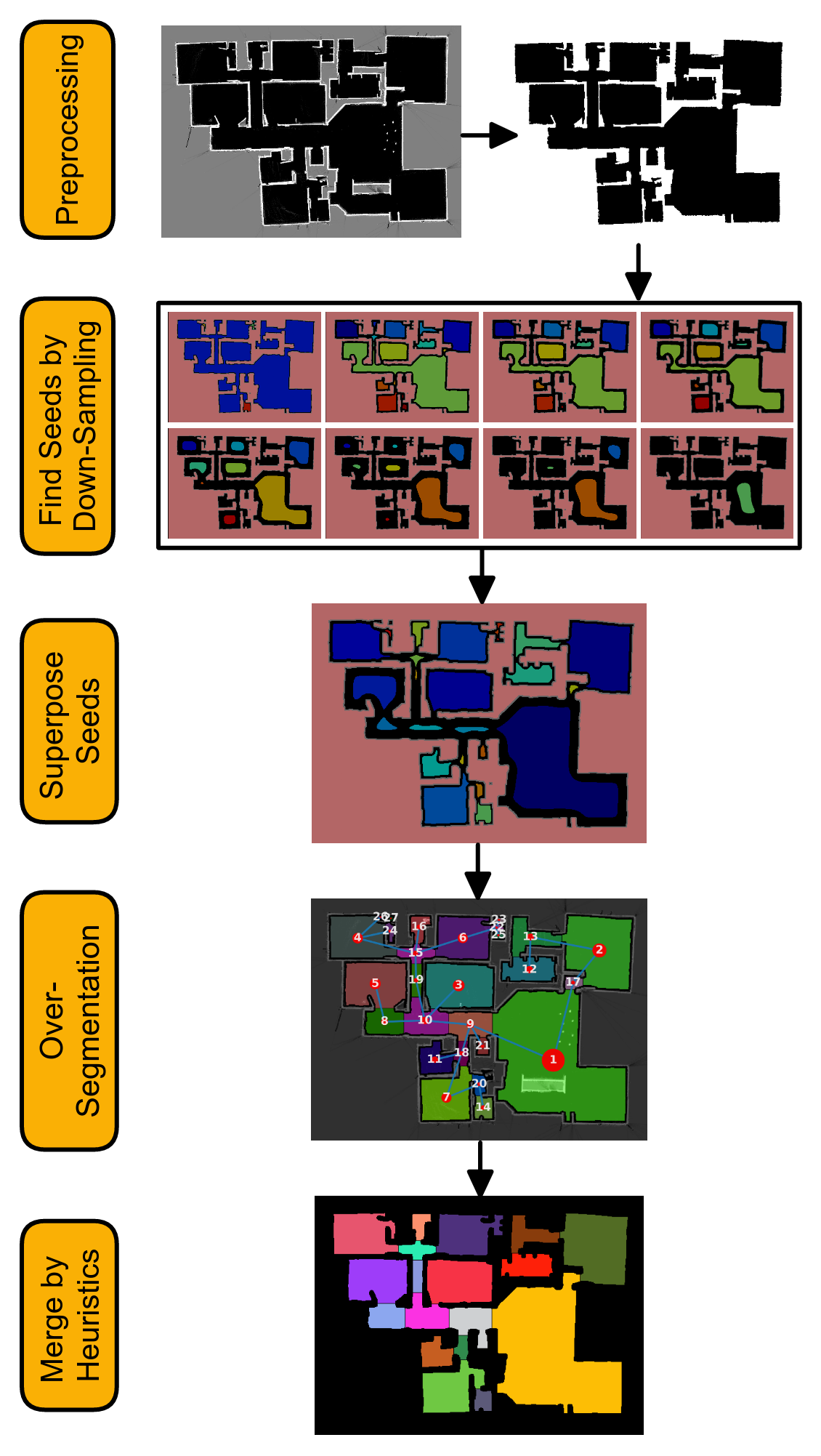}
        \caption{The pipeline of our proposed down-sampling-based 2D floor plan segmentation.}
        \label{fig:cover}
\end{figure}

This paper focuses on 2D occupancy grids \cite{Moravec1985-zy} (a grid map or simply a map) generated by mobile robots with 2D laser range scanners performing Simultaneous Localization and Mapping (SLAM) in an environment. Our algorithm should perform segmentation despite imperfections caused by sensor noise, localization errors, furniture, and other mapping artifacts. We propose an algorithm based on iterative down-sampling of the map to extract core segment information (i.e., seeds). Our algorithm employs classic computer vision techniques and is simple to implement. We utilize the seeds as initial markers for the Watershed segmentation and apply three heuristic merging rules to obtain the final segmentation map. Finally, we discuss well-known measures to quantify the segmentation algorithm's performance. The contributions of this paper are as follows:

\begin{enumerate}
        \item A novel and automatic method for fast and robust segmentation of cluttered 2D floor plans based on down-sampling.
        \item An open-source implementation of the proposed algorithm\footnote{\url{https://github.com/sharif1093/py_floor_plan_segmenter}}.
        \item An evaluation of existing performance metrics for floor plan segmentation and suggestions for future evaluations.
\end{enumerate}

\section{RELATED WORKS}

Bormann et al. \cite{Bormann2016-ze} discussed four major state-of-the-art methods for 2D floor plan segmentation. Morphological segmentation \cite{Buschka2002-sx} uses morphologic operations to find partial segments, and then completes them using the wavefront algorithm. The distance-transform-based segmentation \cite{Bormann2016-ze} iteratively searches over a range of thresholds to find room centers, selecting the threshold that maximizes the number of rooms. The Voronoi-graph-based segmentation \cite{Thrun1998-yt} utilizes a Voronoi graph of the floor plan and identifies the shared boundaries of segments as critical points. Heuristic merging rules \cite{Bormann2016-ze} are then used to improve the segmentation quality. Feature-based segmentation \cite{Martinez_Mozos2006-al} trains a classifier that classifies each pixel based on features visible from that pixel. Among the four segmentation algorithms, Voronoi-graph-based segmentation provides the best performance based on the average recall and precision measure, while morphological and distance-transform-based segmentation are the most computationally efficient methods. Feature-based segmentation has a long runtime and is still inferior in terms of performance compared to the Voronoi-graph-based segmentation.

In their recent work, Luperto et al. \cite{Luperto2022-xw} proposed a structure detection method for identifying dominant angles in a floor plan using Discrete Fourier Transform (DFT), and then utilized this information to declutter the map and perform segmentation. While this technique is intriguing and can also be used for predicting unobserved areas in the map, it assumes the existence of dominant angles in the floor plan, which may not be present in certain structures. Additionally, in highly cluttered environments, this method tends to overlook the clutter in the segmentation, which may be useful in certain scenarios but not necessarily in all.

There are also some studies that use 3D point clouds as input but focus on a projected 2D map and 2D floor plan segmentation. Bobkov et al. \cite{Bobkov2017-xg} utilize clustering on potential fields to detect room boundaries. Turner et al. \cite{Turner2014-ev} use a triangulation of the floor plan and graph min-cut to minimize inter-room boundaries. Ambrus et al. \cite{Ambrus2017-zj} employ an energy minimization algorithm solved by graph-cut to partition the rooms. However, their method relies on the successful detection of different features such as walls, ceilings, and openings. While there are other works \cite{Armeni2016-rf} that use 3D point clouds as input, since they leverage the 3D aspects of the environment, they are beyond the scope of this paper.

Another line of research focuses on parsing architectural floor plan layouts for reconstruction or analysis purposes. These works typically aim to read and parse symbols such as windows, doors, and walls from the floor plan, and then either reconstruct \cite{Lv2021-ej} or vectorize the map \cite{Liu2017-nj}. Some works still address 2D floor plan segmentation, such as Mace et al. \cite{Mace2010-vq}, which detects convex rooms by solving the associated polygon partitioning problem.

To facilitate machine-learning-based 2D floor plan segmentation methods, Li et al. \cite{Li2020-la} have proposed a method to generate synthetic 2D floor plans, and a PseudoSLAM technique to quickly create GMapping-like floor plans for training purposes. Meanwhile, Cruz et al. \cite{Cruz2021-to} provide a 3D dataset of indoor environments. Although there are more learning-based methods in the literature \cite{Kaleci2022-ax, Fleer2017-pi}, they all share the same limitation of being only effective on the data they were trained on. In contrast, our proposed method does not rely on any machine learning, making it applicable to new scenarios without the need for additional training.

\section{METHODS}

Assuming an occupancy grid $O=[C_{ij}]$, where $C_{ij}$ is equal to $0$ when the cell is empty with $100\%$ probability and $255$ when the cell is occupied with $100\%$ probability, the 2D floor plan segmentation problem involves assigning a label to every \textit{empty} cell. Our segmentation method consists of four stages: 1) preprocessing, which removes noise resulting from furniture or mapping artifacts in the input map, 2) seeding, which identifies all possible seeds for potential segments using a down-sampling method, 3) over-segmentation, which uses the superposed seed map and performs Watershed segmentation to produce an over-segmentation, and 4) merging, which forms the segment connectivity graph and performs merges based on simple heuristics. Our focus in this paper is on segmenting the cells rather than obtaining the underlying floor plan layout, as this information is more useful to a mobile robot navigating a cluttered environment. In this case, only the accessible cells matter to the robot, rather than the hypothetical underlying floor plan.

Our proposed method bears similarities to morphological segmentation methods \cite{Buschka2002-sx, Bormann2015-hs} in terms of its implementation. However, our method uses a Gaussian smoothing filter and down-sampling instead of morphological operations. We demonstrate that our method's performance is comparable to or surpasses that of Voronoi-graph-based segmentation while being computationally less expensive. Furthermore, our approach does not rely on learning from data, making it readily usable for new scenarios. We provide a detailed explanation of each sub-step below.

\subsection{Preprocessing}

The goal of the preprocessing step is to improve the robustness of the segmentation results by removing noise and clutter from the map. The first stage of preprocessing involves applying standard denoising techniques, including binary thresholding and removal of small connected components in the foreground and background. Next, the map is aligned so that its dominant directions are horizontal or vertical. To achieve this, we use a probabilistic Hough transform to find line segments in the map, and then create a histogram of line directions with each line's length serving as its weight. The direction bin with the highest weight is used to rotate the map.

\begin{figure}[tpb]
	\centering
        \includegraphics[width=0.48\textwidth]{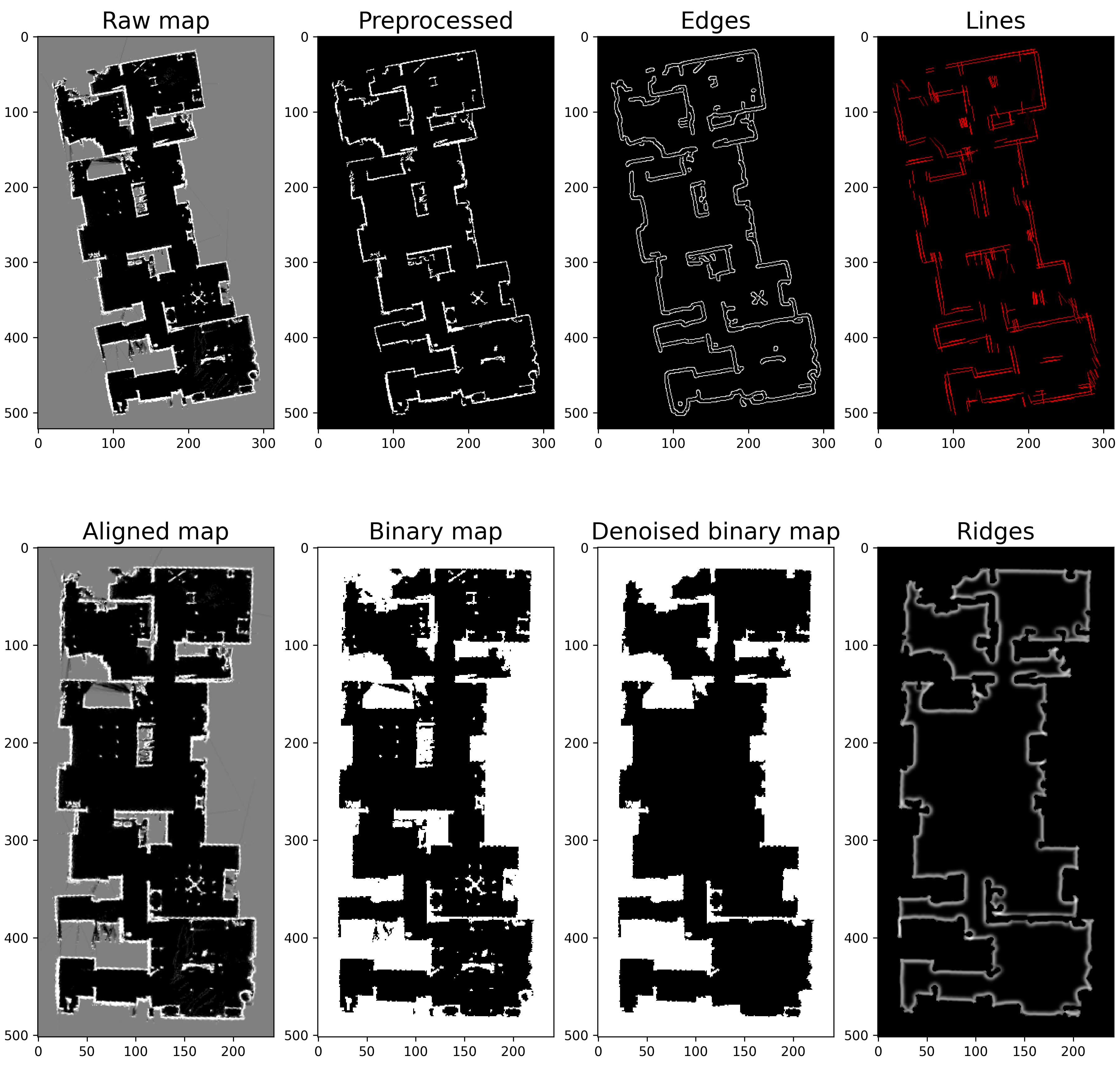}
        \caption{Preprocessing steps prior to segmentation. Top row, from left to right: (1) raw occupancy grid map, (2) map after preprocessing for alignment, (3) Canny edge detection output, and (4) line segments detected by probabilistic Hough transform. Bottom row, from left to right: (5) aligned map after rotation, (6) binary map after thresholding, (7) denoised binary map with small connected components removed, and (8) ridges detected using the Meijering filter.}
        \label{fig:preprocessing}
\end{figure}

After map alignment, we apply the Meijering ridge operator \cite{Meijering2004-lr}, a second-order Hessian-based operator that enhances edges and suppresses other features in an image. This step helps to further improve the segmentation results by enhancing the edges of the map. The entire preprocessing pipeline is illustrated in \autoref{fig:preprocessing}, which shows the output of each step for a typical map. We note that this preprocessing step is crucial for improving the robustness of our method to noise and clutter in the map.

\subsection{Iterative Down-sampling for Seeding}

\begin{figure}[tpb]
        \centering
	\begin{tabular}[c]{cccc}
		\begin{subfigure}{0.1\textwidth}
			\centering
			\includegraphics[width=\linewidth]{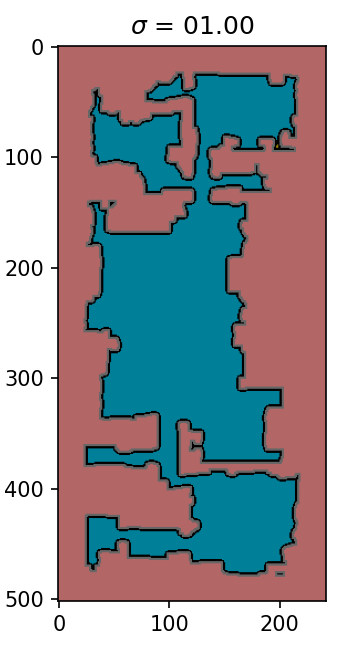}
		\end{subfigure}&
                \begin{subfigure}{0.1\textwidth}
			\centering
			\includegraphics[width=\linewidth]{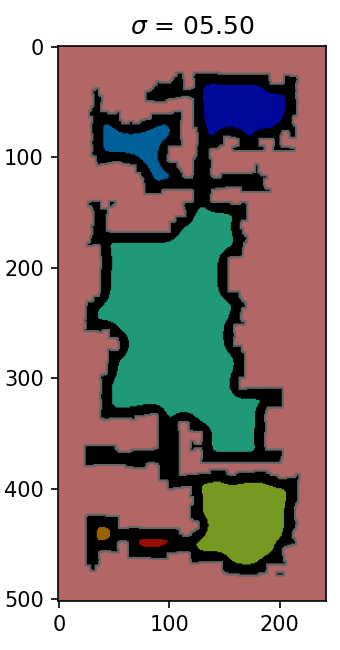}
		\end{subfigure}&
                \begin{subfigure}{0.1\textwidth}
			\centering
			\includegraphics[width=\linewidth]{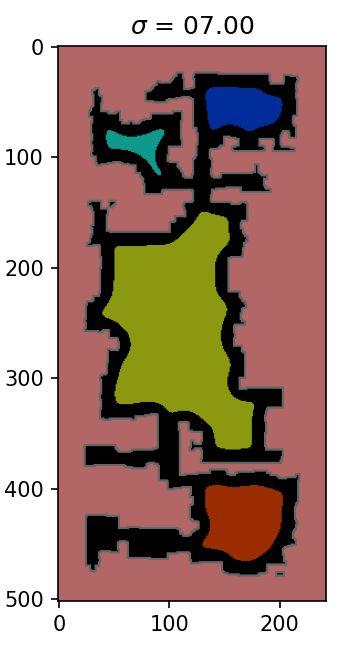}
		\end{subfigure}&
		\begin{subfigure}{0.1\textwidth}
			\centering
			\includegraphics[width=\linewidth]{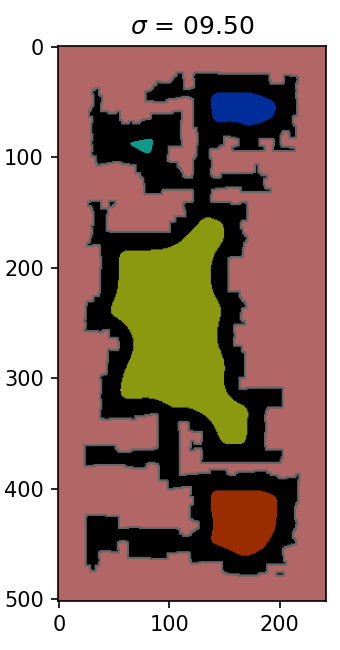}
		\end{subfigure}\\
                \begin{subfigure}{0.1\textwidth}
			\centering
			\includegraphics[width=\linewidth]{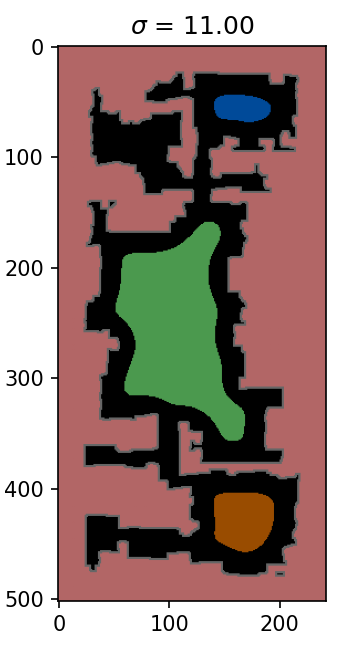}
		\end{subfigure}&
                \begin{subfigure}{0.1\textwidth}
			\centering
			\includegraphics[width=\linewidth]{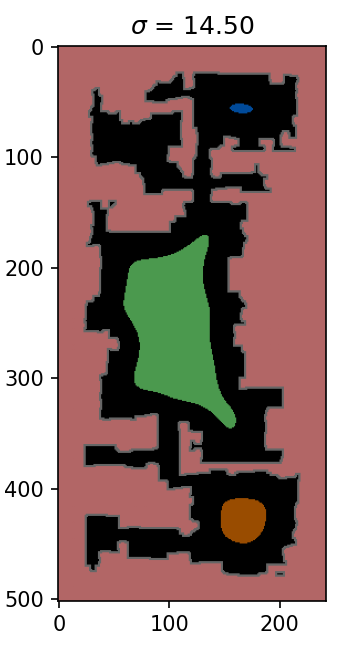}
		\end{subfigure}&
                \begin{subfigure}{0.1\textwidth}
			\centering
			\includegraphics[width=\linewidth]{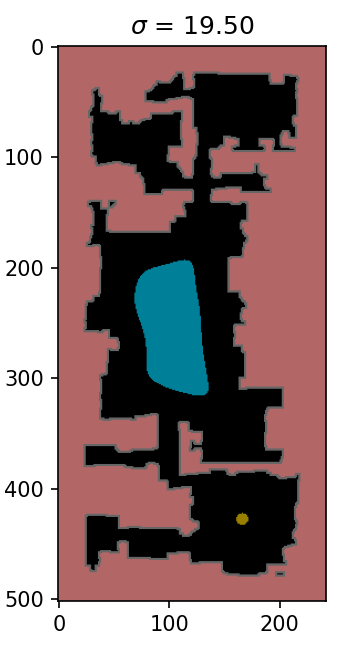}
		\end{subfigure}&
		\begin{subfigure}{0.1\textwidth}
			\centering
			\includegraphics[width=\linewidth]{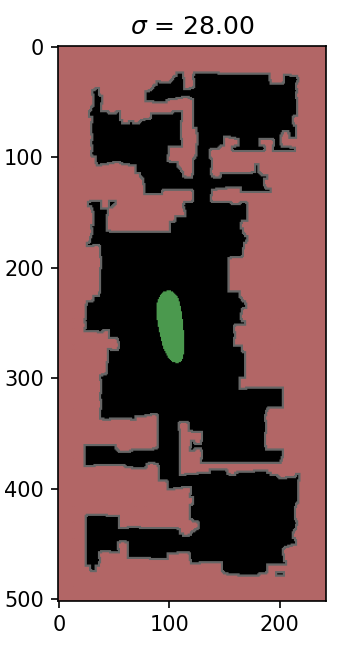}
		\end{subfigure}\\
	\end{tabular}
        
        \caption{The thresholded Gaussian filter results for eight different values of $\sigma$, demonstrating the iterative incrementation of $\sigma$ and the reduction of seeds (i.e., disjoint areas) in the map as a result.}
        \label{fig:seeding}
\end{figure}

In this step, we use an iterative Gaussian blur filter on the map to down-sample the image by removing high-frequency features. The Gaussian blur filter is a separable filter that can be applied as two consecutive 1D filters on the image. The filter's complexity is of order $\mathcal{O}(kWH)$, where $k$ is the Gaussian kernel size linearly proportional to $\sigma$, $W$ is the image width, and $H$ is its height. After applying the filter, we threshold the image using a threshold of $0.95$. We iteratively increase the strength $\sigma$ of the Gaussian blur filter from an initial value of $\sigma_0=1$ until there are no features remaining in the map. We add an eroded version of the binary map to the filtered map itself to represent the background area. Each disjoint area in the map is labeled using a distinct integer value ranging from $0$ for the background area to the maximum number of disjoint areas in each step. We call the labeled filtered map at iteration $i$ a \textit{seed} map, $\mathcal{S}_i$, which is stored in the set $\mathcal{S}=[\mathcal{S}_i]$. An illustration of this process is shown in \autoref{fig:seeding}.

\subsection{Over Segmentation}

Starting from multiple disjoint seeds at iteration $i=0$, consisting of a background and foreground seeds, the foreground seeds constantly diminish and may divide into two or more seeds as sigma increases. Our aim is to create a superposed map of all seeds that did not divide after their creation, referred to as leaf seeds in the seed \textit{graph}. To track the birth and death of each seed, we begin by reversing from the last seed map, $\mathcal{S}_N$, where $N$ is the number of seed maps in $\mathcal{S}$. At each frame index $i$, we create a local mapping, $\mathcal{L}_i^{i+1}$, from seeds in $\mathcal{S}_i$ to seeds in $\mathcal{S}_{i+1}$. We define $\Gamma$ as an operator that gives the set of all distinct labels in a seed map. For a label $l_j\in\Gamma(\mathcal{S}_i)$, we consider three cases regarding all regions labeled as $l_j$ at iteration $i+1$: (a) there are no seeds in $\Gamma(\mathcal{S}_{i+1})$, indicating that $i$ was the death moment for the seed corresponding to $l_j$; (b) there is exactly one label in $\Gamma(\mathcal{S}_{i+1})$, implying that $i$ is a normal moment during the life of the seed corresponding to $l_j$; and (c) there are more than one label in $\Gamma(\mathcal{S}_{i+1})$, indicating that $i$ is the moment when seed $l_j$ is dividing into two or more child seeds in $i+1$, and hence, it is the birth moment for the child seed. This algorithm identifies the child seeds as well as their birth and death moments. Finally, we create a superposed seed map, $\bar{\mathcal{S}}$, of all leaf seeds exactly after their birth, when they were created from a parent seed's division.

\begin{figure}[tpb]
	\centering
        \includegraphics[width=0.48\textwidth]{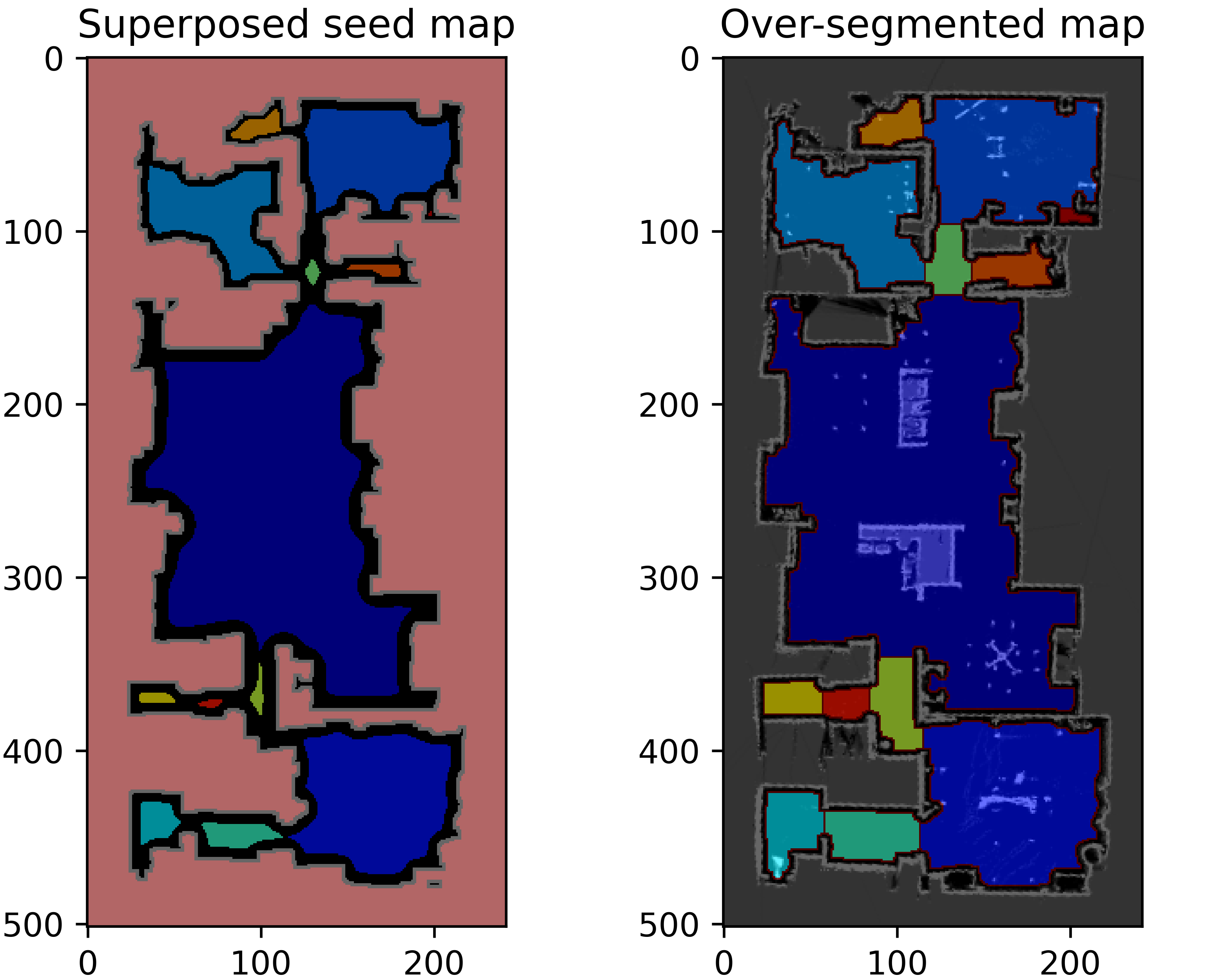}
        \caption{On the left, the superposed seed map. On the right, the Watershed segmentation of the map using the superposed seed map as the initial markers, resulting in an over-segmented map.}
        \label{fig:oversegmented}
\end{figure}

After obtaining the superposed seed map, we use it as the initial markers for the Watershed segmentation. To segment only the foreground cells of the map, we utilize the Meijering \cite{Meijering2004-lr} ridge detector and perform Watershed on the ridge map. Ridges exhibit high intensity and their steep gradient guides the Watershed segmentation to respect the ridges as the boundaries of the segmented maps. This produces an over-segmented map, denoted as $O$, as shown in \autoref{fig:oversegmented}.

\subsection{Merging}

In order to perform merging, we first represent the segment connectivity graph as an attributed graph $G=(V,E)$, where each node in $V$ represents a segment in the over-segmented map $O$, and each edge in $E$ indicates whether a shared boundary exists between two nodes. Each node of the graph stores the area and the centroid of the segment it represents. Each edge of the graph stores the length of the shared boundary of the segments it connects. We define the following simple heuristics for merging the nodes:

\begin{enumerate}
        \item If a node has no edges, it will be removed from the graph if its area is below a threshold, $A_0=\max(1\text{ m}^2, 0.01 \times A_{\text{total}})$, where $A_{\text{total}}$ is the total area of the floor plan.
        \item Starting from leaf nodes, they will be merged to their neighbor that has the most shared boundary with them if the node's area is below the area threshold $A_0$.
        \item Starting from edges with the highest lengths, nodes will be merged if their edge length (shared boundary) is beyond a threshold $L_0=3\text{ m}$.
\end{enumerate}

\begin{figure}[tpb]
        \centering
	\begin{tabular}[c]{cccc}
		\begin{subfigure}{0.175\textwidth}
			\centering
			\includegraphics[width=\linewidth]{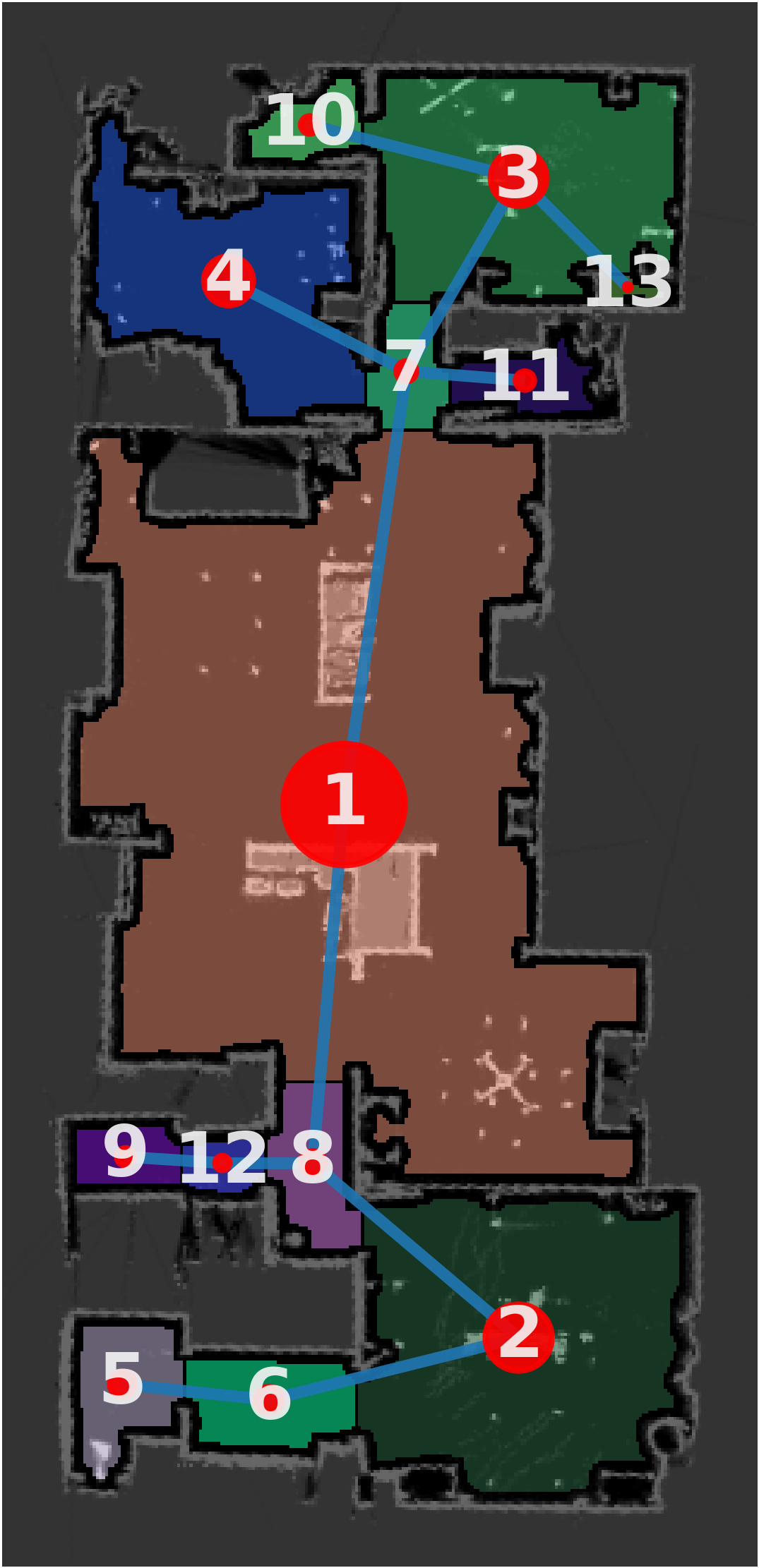}
                        \caption{}
		\end{subfigure}&
                \begin{subfigure}{0.175\textwidth}
			\centering
			\includegraphics[width=\linewidth]{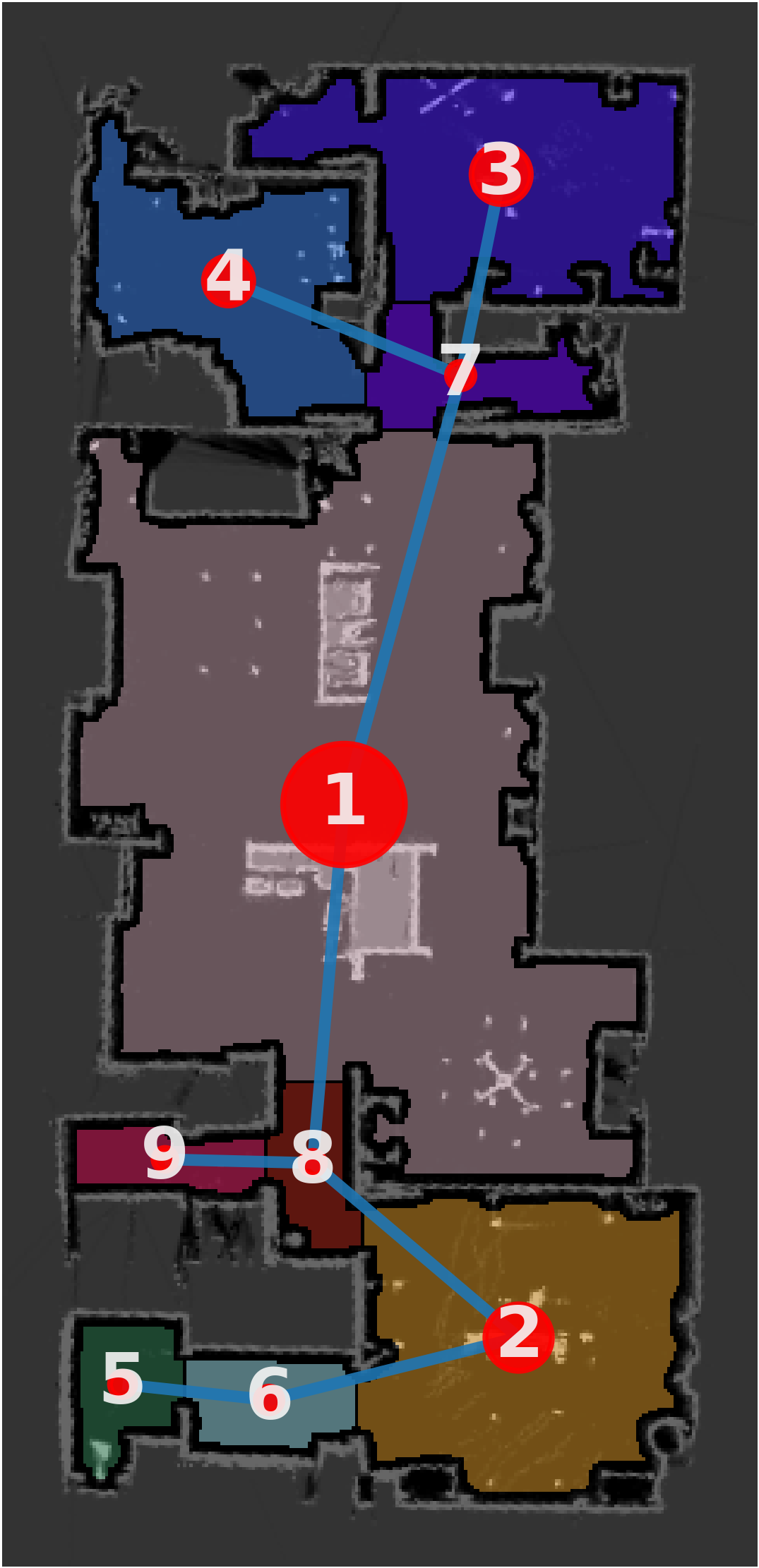}
                        \caption{}
		\end{subfigure}\\
	\end{tabular}
        
        \caption{The overlaid segment connectivity graph (a) before and (b) after applying merge heuristics.}
        \label{fig:merge}
\end{figure}

The overlaid segmentation map before and after merge is shown in \autoref{fig:merge}.

\section{EVALUATION \& RESULTS}

\begin{figure*}[!htb]
	\centering
        \includegraphics[width=0.98\textwidth]{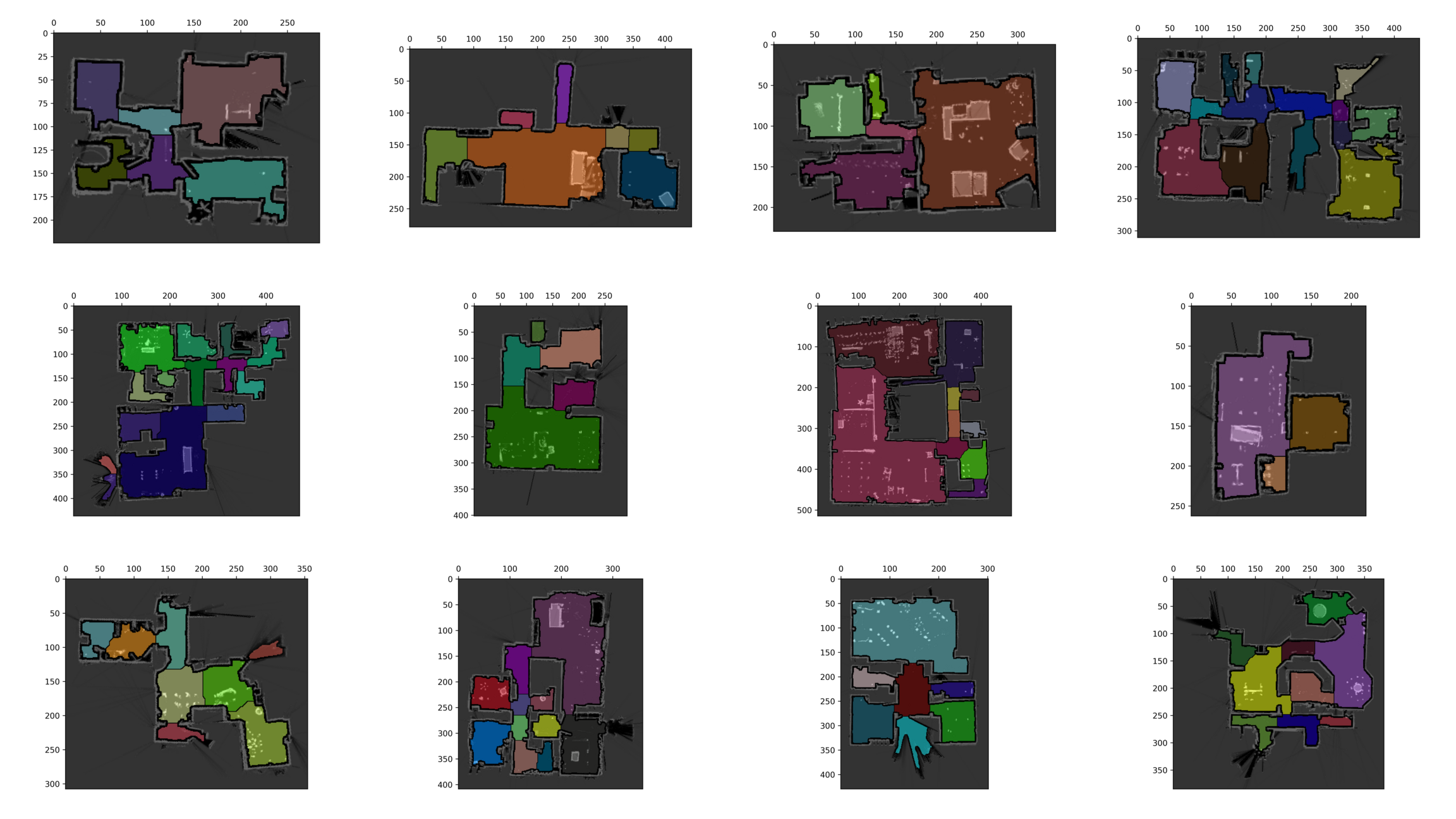}
        \caption{The segmentation results are presented for 12 real and cluttered floor plans, which were obtained using Neato vacuum cleaning robots.}
        \label{fig:neato}
\end{figure*}

First, we evaluate the performance of the proposed down-sampling-based 2D floor plan segmentation method on a batch of sample maps generated by Neato D-series vacuum cleaning robots across the US and Europe in anonymized sampled floor plans. The algorithm is independent of the method used for map generation, such as the SLAM algorithm. The robot uses a 2D laser scanner that scans the environment at a rate of $4Hz$. Since the environment is unknown, there is no ground truth for the sampled maps, and only a qualitative evaluation is performed. The maps are highly cluttered due to furniture, noise, and mapping artifacts and are fed to the algorithm as-is. The results are shown in \autoref{fig:neato}, where the scales of the maps are shown on the images, and each unit is equal to $4cm$. As seen in the figure, our method exhibits great resilience towards clutter in the environment. Moreover, it is evident that the proposed method performs semantically meaningful segmentation despite only using a 2D map. The algorithm does not make any assumptions about the floor plan's structure, i.e., non-Manhattan worlds are segmented successfully, as seen in the bottom right-most map in \autoref{fig:neato}. However, in some cases, an over-segmentation of the map is noted, typically for long hallways, as in the top-most right-most map in \autoref{fig:neato}.

\begin{figure*}[!htb]
        \centering
	\begin{tabular}[c]{c}
		\begin{subfigure}{0.98\textwidth}
			\centering
			\includegraphics[width=\linewidth]{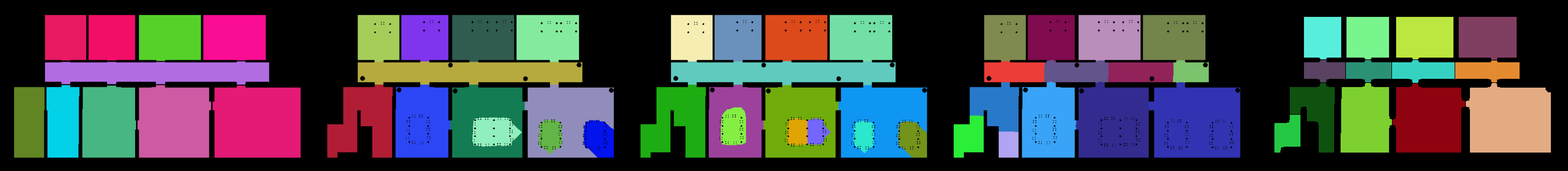}
		\end{subfigure}\\
                \begin{subfigure}{0.98\textwidth}
			\centering
			\includegraphics[width=\linewidth]{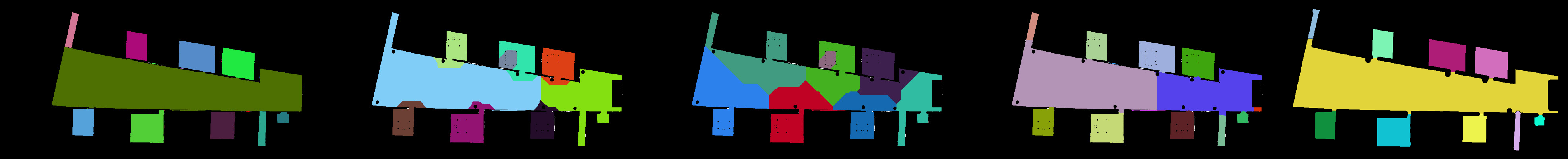}
		\end{subfigure}\\
                \begin{subfigure}{0.98\textwidth}
			\centering
			\includegraphics[width=\linewidth]{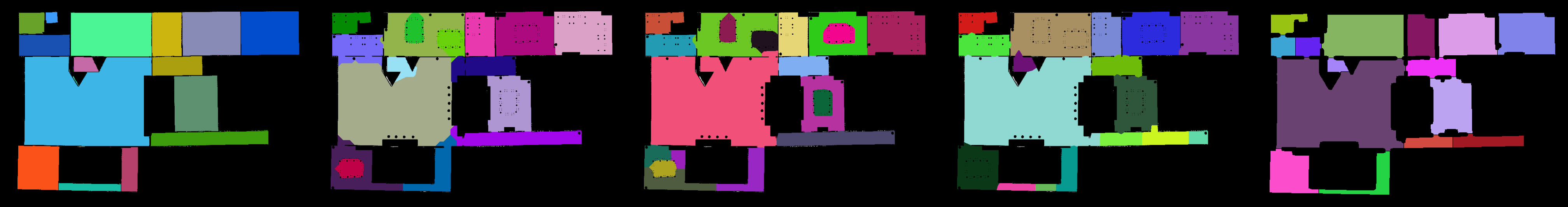}
		\end{subfigure}\\
                \begin{subfigure}{0.98\textwidth}
			\centering
			\includegraphics[width=\linewidth]{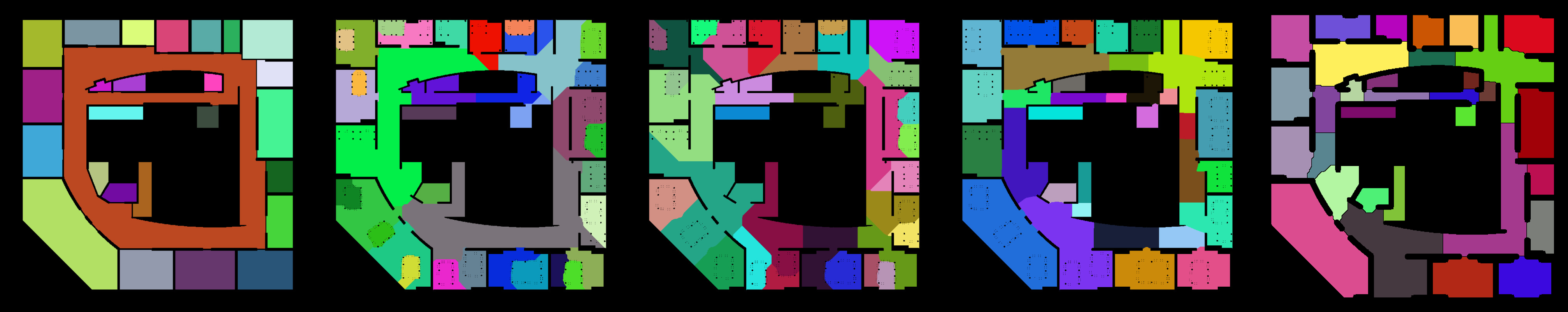}
		\end{subfigure}\\
                \begin{subfigure}{0.98\textwidth}
                        \centering
                        \includegraphics[width=\linewidth]{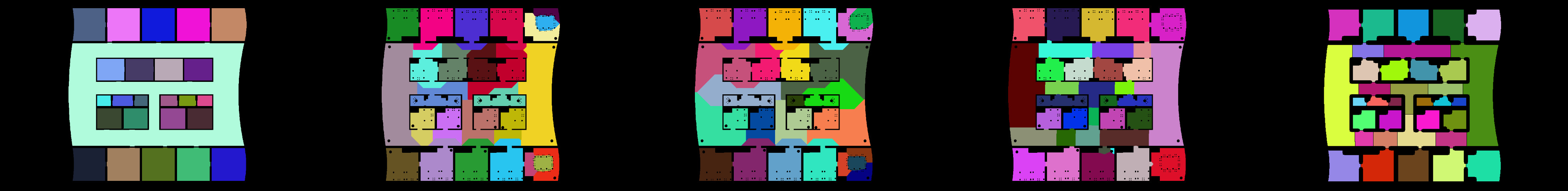}
                \end{subfigure}\\
	\end{tabular}
        
        \caption{Comparison of methods. Left to right: (1) Ground truth, (2) morphologic, (3) distance transform, (4) Voronoi graph, and (5) ours. Map numbers top to bottom: 1, 3, 7, 9, and 15.}
        \label{fig:benchmark}
\end{figure*}

\begin{table*}[t!]
	\centering
	\caption{Performance of the proposed method using different evaluation metrics for each furnished map in the benchmark \cite{Bormann2016-ze}}
	\label{tab:measures}
	\begin{tabular}{lccccccccccc}
		\toprule
		\textbf{Map} & \textbf{$\mathfrak{U}$} & \textbf{Recall} & \textbf{$\mathfrak{C} $} & \textbf{Precision} & \textbf{$\mathfrak{H}$} & \textbf{V-score} & \textbf{AMI} & \textbf{NMI} & \textbf{ARI} & \textbf{FMI} & \textbf{IoU} \\ [0.5ex]
		\midrule
                1  & 11.74\% & 89.96\% & 91.12\% & 98.63\%  & 98.55\%  & 94.69\% & 94.69\% & 94.69\% & 92.42\% & 93.45\% & 88.17\% \\ 
                2  & 19.83\% & 86.99\% & 89.48\% & 100.00\% & 100.00\% & 94.45\% & 94.44\% & 94.45\% & 85.09\% & 86.92\% & 86.99\% \\ 
                3  & 9.51\%  & 98.12\% & 99.03\% & 99.49\%  & 98.49\%  & 98.76\% & 98.76\% & 98.76\% & 99.19\% & 99.59\% & 97.60\% \\ 
                4  & 14.67\% & 97.67\% & 66.28\% & 88.02\%  & 94.23\%  & 77.82\% & 77.77\% & 77.82\% & 21.41\% & 37.17\% & 70.59\% \\ 
                5  & 13.79\% & 91.35\% & 96.68\% & 97.25\%  & 97.56\%  & 97.12\% & 97.11\% & 97.12\% & 96.84\% & 97.04\% & 90.14\% \\
                6  & 18.59\% & 93.52\% & 89.01\% & 95.76\%  & 97.33\%  & 92.98\% & 92.98\% & 92.98\% & 84.58\% & 86.26\% & 91.76\% \\ 
                7  & 13.91\% & 92.80\% & 97.19\% & 97.02\%  & 98.36\%  & 97.77\% & 97.77\% & 97.77\% & 99.10\% & 99.25\% & 80.10\% \\ 
                8  & 16.67\% & 91.13\% & 92.89\% & 96.50\%  & 97.07\%  & 94.94\% & 94.92\% & 94.94\% & 92.65\% & 93.04\% & 88.17\% \\ 
                9  & 12.85\% & 96.84\% & 76.23\% & 95.16\%  & 93.49\%  & 83.98\% & 83.97\% & 83.98\% & 42.54\% & 52.83\% & 82.63\% \\ 
               10  & 18.61\% & 82.83\% & 83.98\% & 96.83\%  & 95.67\%  & 89.44\% & 89.44\% & 89.44\% & 87.69\% & 89.64\% & 80.16\% \\ 
               11  & 14.53\% & 96.83\% & 83.87\% & 98.21\%  & 98.89\%  & 90.76\% & 90.74\% & 90.76\% & 55.02\% & 62.09\% & 93.04\% \\ 
               12  & 12.33\% & 90.72\% & 82.01\% & 99.99\%  & 99.81\%  & 90.04\% & 90.03\% & 90.04\% & 65.27\% & 70.88\% & 90.70\% \\ 
               13  & 14.38\% & 96.89\% & 78.57\% & 100.00\% & 100.00\% & 88.00\% & 87.99\% & 88.00\% & 46.51\% & 56.88\% & 96.89\% \\ 
               14  & 17.72\% & 95.53\% & 91.91\% & 99.93\%  & 99.92\%  & 95.75\% & 95.74\% & 95.75\% & 80.36\% & 82.55\% & 95.53\% \\ 
               15  & 15.04\% & 97.19\% & 73.48\% & 100.00\% & 100.00\% & 84.71\% & 84.70\% & 84.71\% & 35.72\% & 50.86\% & 97.19\% \\ 
               16  & 14.15\% & 95.63\% & 84.35\% & 99.85\%  & 99.86\%  & 91.45\% & 91.44\% & 91.45\% & 54.99\% & 62.92\% & 95.42\% \\ 
               17  & 17.20\% & 96.62\% & 83.99\% & 100.00\% & 100.00\% & 91.30\% & 91.29\% & 91.30\% & 64.62\% & 70.53\% & 96.62\% \\ 
               18  & 10.66\% & 97.53\% & 86.05\% & 99.71\%  & 99.34\%  & 92.22\% & 92.21\% & 92.22\% & 63.64\% & 69.39\% & 97.14\% \\ 
               19  & 13.37\% & 95.66\% & 81.28\% & 99.95\%  & 99.72\%  & 89.56\% & 89.56\% & 89.56\% & 56.96\% & 64.99\% & 95.60\% \\ 
               20  & 8.62\%  & 97.11\% & 92.59\% & 99.96\%  & 99.93\%  & 96.12\% & 96.12\% & 96.12\% & 79.48\% & 81.92\% & 97.11\% \\
               \midrule
               AVE & 14.41\% & 94.05\% & 86.00\% &  98.11\% &  98.41\% & 91.59\% & 91.58\% & 91.59\% & 70.20\% & 75.41\% & 90.58\% \\
               STD &  3.04\% & 4.10\%  & 8.37\%  &  2.88\%  &  1.98\%  &  5.18\% &  5.18\% & 5.18\%  & 22.87\% & 18.13\% & 7.40\%  \\
	       \bottomrule
	\end{tabular}
\end{table*}

\begin{table*}[!t]
	\centering
	\caption{Comparison of our results with those of Bormann et al. \cite{Bormann2016-ze}. Only furnished benchmark maps are considered.}
	\label{tab:comparison}
	\begin{tabular}{cccccc}
		\toprule
		             &           & \textbf{morph}   & \textbf{distance} & \textbf{Voronoi} & \textbf{ours}    \\ [0.5ex]
		\midrule
                no furniture & recall    & 98.1\%$\pm$2.4\% & 96.9\%$\pm$2.8\%  & 95.0\%$\pm$2.3\% & 94.8\%$\pm$3.4\% \\
                             & precision & 88.5\%$\pm$9.2\% & 88.4\%$\pm$9.3\%  & 94.8\%$\pm$5.0\% & 98.2\%$\pm$3.0\% \\
                \midrule
                furnished    & recall    & 84.6\%$\pm$7.2\% & 76.1\%$\pm$12.3\% & 86.6\%$\pm$5.2\% & 94.1\%$\pm$4.1\% \\
                             & precision & 90.5\%$\pm$8.1\% & 88.4\%$\pm$8.5\%  & 94.5\%$\pm$5.1\% & 98.1\%$\pm$2.9\% \\
	        \bottomrule
	\end{tabular}
\end{table*}

In order to compare the results of the proposed method with the state-of-the-art in the literature, we executed the proposed method on the benchmark floor plans provided by Bormann et al. \cite{Bormann2016-ze}. Several samples are shown in \autoref{fig:benchmark}, alongside the top three state-of-the-art methods provided in \cite{Bormann2016-ze}: morphological segmentation, distance-transform-based segmentation, and Voronoi-graph-based segmentation. The complete results are available online\footnote{\url{https://github.com/sharif1093/py_floor_plan_segmenter}}. Upon qualitative comparison, the results of our method are similar to the results from the Voronoi-graph-based segmentation, but with less computational effort and simpler heuristics. Fewer and simpler heuristics suggest that our algorithm would work more readily and out-of-the-box on unseen cases. The average runtime of our proposed method on the 20 samples from the benchmark is $10.67s\pm 20.25s$ and $10.31s\pm 19.1s$ for the unfurnished and furnished maps, respectively. The benchmarking was conducted on a laptop equipped with an Intel® Core™ i7-10875H CPU @ 2.30GHz.

\subsection{Performance Metric}

There is no consensus on the metrics used for evaluating the performance of floor plan segmentation results. For example, Luperto et al. \cite{Luperto2022-xw} use Intersection over Union (IoU), while Bormann et al. \cite{Bormann2016-ze} suggest a specific definition of recall and precision to measure performance. We report some of the commonly used metrics in the literature for the 20 benchmark maps in \autoref{tab:measures}: (1) under-painting, $\mathfrak{U}$, (2) recall, as defined in \cite{Bormann2016-ze}, (3) completeness \cite{Rosenberg2007-qe}, $\mathfrak{C}$, (4) precision, as defined in \cite{Bormann2016-ze}, (5) homogeneity \cite{Rosenberg2007-qe}, $\mathfrak{H}$, (6) V-score \cite{Rosenberg2007-qe}, (7) Adjusted Mutual Information (AMI) \cite{Vinh2010-ti}, (8) Normalized Mutual Information (NMI) \cite{Vinh2010-ti}, (9) Adjusted Rand Index (ARI) \cite{Steinley2004-le}, (10) Fowlkes-Mallows Index (FMI) \cite{Fowlkes1983-an}, and (11) Intersection over Union (IoU). The under-painting ($\mathfrak{U}$) is defined as the number of cells that have not received any labels but have a label in the ground truth, divided by the overall number of non-background ground truth pixels. A lower value is better. All metrics in the table are calculated after the under-painted pixels are subtracted from the set of predicted and ground truth labels to itemize the performance and identify the reason for failure. We include the average precision and recall for the proposed method for both furnished and unfurnished benchmark maps in \autoref{tab:comparison}. We also include the results from Bormann et al. \cite{Bormann2015-hs}, although this comparison might not be fair because the under-painted cells are included in the results of Bormann et al., but not in ours.

Let us define a \textit{class} as the set of ground-truth labels and a \textit{cluster} as the set of predicted labels. Recall intuitively measures how well an algorithm is connecting predicted segments to avoid over-segmentation, i.e., how focused the distribution of classes is over clusters. On the other hand, precision measures the algorithm's ability to not mix different ground-truth segments into a predicted segment, i.e., how focused the distribution of clusters is over classes.

Mathematically, we can represent the contingency matrix as $A=[a_{ck}]$, where $c$ represents a class, $k$ represents a cluster, and $a_{ck}$ denotes the frequency of items assigned to cluster $k$ and belonging to class $c$. Then, recall and precision are defined as follows:

\begin{align}
        r &= \sum_{c=1}^{|C|}{\left[ \frac{1}{|C|} \frac{\max_{k}{a_{ck}}}{\sum_{k=1}^{|K|}{a_{ck}}} \right]} \\
        p &= \sum_{k=1}^{|K|}{\left[ \frac{1}{|K|} \frac{\max_{c}{a_{ck}}}{\sum_{c=1}^{|C|}{a_{ck}}} \right]},
\end{align}

\noindent where $K$ and $C$ denote the set of clusters and classes, respectively. Completeness and homogeneity are other metrics that almost measure the same concept, which are the extent to which each cluster contains items of a single class and each class is assigned to a single cluster, respectively. Formally, they are defined as:

\begin{align}
        \mathfrak{C} &= 1 - \frac{H(K|C)}{H(K)} \\
        \mathfrak{H} &= 1 - \frac{H(C|K)}{H(C)}
\end{align}

\noindent where $H(K|C)$ and $H(C|K)$ are the conditional entropies of either $K$ or $C$ given the other, and $H(K)$ and $H(C)$ are the entropies of the clusters and classes, respectively. 

As noted, recall is equivalent to completeness while precision is equivalent to homogeneity. For example, in cases of pure over-segmentation, each class may have all of the predicted clusters, making knowledge of the class irrelevant in knowing the cluster. This results in $H(K|C)$ being maximal and equal to $H(K)$, which in turn means $\mathfrak{C}=0$. Our results in the table generally confirm this agreement between the metrics. However, there are cases where these measures do not exactly agree. For instance, in map 15 in \autoref{tab:measures}, recall=$97.19\%$ while $\mathfrak{C}=73.48\%$, and IoU=$97.19\%$, suggesting good segmentation results. The segmentation results for map 15 are included in the last row of \autoref{fig:benchmark}. However, there is an over-segmentation in the map, especially in the corridor segment. Recall and IoU are both insensitive to this because they average the classes without considering their size. Depending on the application, completeness and homogeneity may be more suitable measures compared to recall and precision in capturing the performance of the segmentation.

If a single metric is required to evaluate the segmentation performance, V-measure \cite{Rosenberg2007-qe}, which is the harmonic mean of completeness and homogeneity, can be used. As evident from \autoref{tab:measures}, AMI, V-score, and NMI have a high level of agreement, and thus any of them can be used empirically as a single measure of performance. However, ARI and FMI are not suitable for measuring the performance of segmentation when classes are highly unbalanced \cite{Romano2016-mn}, which may often be the case for most floor plan segmentation problems. IoU, similar to recall and precision, does not consider the size of the class, and thus has the same issue. Overall, measuring performance is highly dependent on the application.

The combination of homogeneity and completeness, along with either V-score, AMI, or NMI, seem to be appropriate measures for evaluating the performance of floor plan segmentation. Additionally, the concept of under-painting can be useful in distinguishing between failure cases caused by failing to label a pixel versus mislabeling a labeled pixel. This concept is particularly valuable in applications where under-painting is not a significant concern, such as in mobile robot applications where labeling occupied pixels is not a top priority.

\section{CONCLUSIONS}

In this paper, we present a 2D floor plan segmentation method that involves iteratively down-sampling the map using a Gaussian filter, followed by Watershed segmentation to create an over-segmented map based on the superposed seeds as the initial markers. We tested the algorithm on cluttered floor plans obtained from Neato vacuum cleaning robots and benchmark maps from \cite{Bormann2016-ze}. Our results demonstrate that the proposed algorithm performs well in both cluttered and uncluttered environments, with a lower computational cost than Voronoi-graph-based segmentation methods. We also discuss various metrics commonly used for evaluating floor plan segmentation and suggest that under-painting, completeness, and homogeneity may be better measures than recall and precision. Overall, we recommend V-measure or AMI as better overall measures than IoU for evaluating floor plan segmentation performance.

This work has some limitations that can be addressed in future research. One limitation is that the preprocessing stage is limited to pure morphological operations, which may cause the removal of important components from the map. This can be improved by implementing structure extraction methods, such as the one proposed in \cite{Luperto2019-pb}. In addition, future research could explore the application of the down-sampling and hierarchical segmentation approach to 3D point clouds for floor plan segmentation.

\addtolength{\textheight}{-12cm}   



\bibliographystyle{IEEEtran}
\bibliography{IEEEabrv,references}

\begin{thebibliography}{10}
\providecommand{\url}[1]{#1}
\csname url@rmstyle\endcsname
\providecommand{\newblock}{\relax}
\providecommand{\bibinfo}[2]{#2}
\providecommand\BIBentrySTDinterwordspacing{\spaceskip=0pt\relax}
\providecommand\BIBentryALTinterwordstretchfactor{4}
\providecommand\BIBentryALTinterwordspacing{\spaceskip=\fontdimen2\font plus
\BIBentryALTinterwordstretchfactor\fontdimen3\font minus
  \fontdimen4\font\relax}
\providecommand\BIBforeignlanguage[2]{{%
\expandafter\ifx\csname l@#1\endcsname\relax
\typeout{** WARNING: IEEEtran.bst: No hyphenation pattern has been}%
\typeout{** loaded for the language `#1'. Using the pattern for}%
\typeout{** the default language instead.}%
\else
\language=\csname l@#1\endcsname
\fi
#2}}

\bibitem{Turner2014-ev}
E.~Turner and A.~Zakhor, ``Floor plan generation and room labeling of indoor
  environments from laser range data,'' in \emph{2014 International Conference
  on Computer Graphics Theory and Applications ({GRAPP})}, Jan. 2014, pp.
  1--12.

\bibitem{Ambrus2017-zj}
R.~Ambru{\c s}, S.~Claici, and A.~Wendt, ``Automatic room segmentation from
  unstructured {3-D} data of indoor environments,'' \emph{IEEE Robotics and
  Automation Letters}, vol.~2, no.~2, pp. 749--756, Apr. 2017.

\bibitem{Moravec1985-zy}
H.~Moravec and A.~Elfes, ``High resolution maps from wide angle sonar,'' in
  \emph{Proceedings. 1985 {IEEE} International Conference on Robotics and
  Automation}, vol.~2, Mar. 1985, pp. 116--121.

\bibitem{Bormann2016-ze}
R.~Bormann, F.~Jordan, W.~Li, J.~Hampp, and M.~H{\"a}gele, ``Room segmentation:
  Survey, implementation, and analysis,'' in \emph{2016 {IEEE} International
  Conference on Robotics and Automation ({ICRA})}, May 2016, pp. 1019--1026.

\bibitem{Buschka2002-sx}
P.~Buschka and A.~Saffiotti, ``A virtual sensor for room detection,'' in
  \emph{{IEEE/RSJ} International Conference on Intelligent Robots and Systems},
  vol.~1, Sept. 2002, pp. 637--642 vol.1.

\bibitem{Thrun1998-yt}
S.~Thrun, ``Learning metric-topological maps for indoor mobile robot
  navigation,'' \emph{Artif. Intell.}, vol.~99, no.~1, pp. 21--71, Feb. 1998.

\bibitem{Martinez_Mozos2006-al}
O.~Martinez~Mozos, A.~Rottmann, R.~Triebel, P.~Jensfelt, and W.~Burgard,
  ``\BIBforeignlanguage{en}{Semantic labeling of places using information
  extracted from laser and vision sensor data},'' Beijing, China, Oct. 2006.

\bibitem{Luperto2022-xw}
M.~Luperto, T.~P. Kucner, A.~Tassi, M.~Magnusson, and F.~Amigoni, ``Robust
  structure identification and room segmentation of cluttered indoor
  environments from occupancy grid maps,'' \emph{IEEE Robotics and Automation
  Letters}, vol.~7, no.~3, pp. 7974--7981, July 2022.

\bibitem{Bobkov2017-xg}
D.~Bobkov, M.~Kiechle, S.~Hilsenbeck, and E.~Steinbach, ``Room segmentation in
  {3D} point clouds using anisotropic potential fields,'' in \emph{2017 {IEEE}
  International Conference on Multimedia and Expo ({ICME})}, July 2017, pp.
  727--732.

\bibitem{Armeni2016-rf}
I.~Armeni, O.~Sener, A.~R. Zamir, H.~Jiang, I.~Brilakis, M.~Fischer, and
  S.~Savarese, ``{3D} semantic parsing of {Large-Scale} indoor spaces,'' in
  \emph{2016 {IEEE} Conference on Computer Vision and Pattern Recognition
  ({CVPR})}, June 2016, pp. 1534--1543.

\bibitem{Lv2021-ej}
X.~Lv, S.~Zhao, X.~Yu, and B.~Zhao, ``Residential floor plan recognition and
  reconstruction,'' in \emph{2021 {IEEE/CVF} Conference on Computer Vision and
  Pattern Recognition ({CVPR})}.\hskip 1em plus 0.5em minus 0.4em\relax IEEE,
  June 2021.

\bibitem{Liu2017-nj}
C.~Liu, J.~Wu, P.~Kohli, and Y.~Furukawa, ``Raster-to-vector: Revisiting
  floorplan transformation,'' in \emph{2017 {IEEE} International Conference on
  Computer Vision ({ICCV})}.\hskip 1em plus 0.5em minus 0.4em\relax IEEE, Oct.
  2017.

\bibitem{Mace2010-vq}
S.~Mac{\'e}, H.~Locteau, E.~Valveny, and S.~Tabbone, ``A system to detect rooms
  in architectural floor plan images,'' in \emph{Proceedings of the 9th {IAPR}
  International Workshop on Document Analysis Systems}, ser. DAS '10.\hskip 1em
  plus 0.5em minus 0.4em\relax New York, NY, USA: Association for Computing
  Machinery, June 2010, pp. 167--174.

\bibitem{Li2020-la}
T.~Li, D.~Ho, C.~Li, D.~Zhu, C.~Wang, and M.~Q.-H. Meng, ``{HouseExpo}: A
  large-scale {2D} indoor layout dataset for learning-based algorithms on
  mobile robots,'' in \emph{2020 {IEEE/RSJ} International Conference on
  Intelligent Robots and Systems ({IROS})}, Oct. 2020, pp. 5839--5846.

\bibitem{Cruz2021-to}
S.~Cruz, W.~Hutchcroft, Y.~Li, N.~Khosravan, I.~Boyadzhiev, and S.~B. Kang,
  ``\BIBforeignlanguage{en}{Zillow indoor dataset: Annotated floor plans with
  360° panoramas and {3D} room layouts},'' in
  \emph{\BIBforeignlanguage{en}{2021 {IEEE/CVF} Conference on Computer Vision
  and Pattern Recognition ({CVPR})}}.\hskip 1em plus 0.5em minus 0.4em\relax
  IEEE, June 2021.

\bibitem{Kaleci2022-ax}
B.~Kaleci, K.~Turgut, and H.~Dutagaci, ``{2DLaserNet}: A deep learning
  architecture on {2D} laser scans for semantic classification of mobile robot
  locations,'' \emph{Engineering Science and Technology, an International
  Journal}, vol.~28, p. 101027, Apr. 2022.

\bibitem{Fleer2017-pi}
D.~Fleer, ``\BIBforeignlanguage{en}{{Human-Like} room segmentation for domestic
  cleaning robots},'' \emph{\BIBforeignlanguage{en}{Robotics}}, vol.~6, no.~4,
  p.~35, Nov. 2017.

\bibitem{Bormann2015-hs}
R.~Bormann, J.~Hampp, and M.~Hagele, ``\BIBforeignlanguage{en}{New brooms sweep
  clean - an autonomous robotic cleaning assistant for professional office
  cleaning},'' in \emph{\BIBforeignlanguage{en}{2015 {IEEE} International
  Conference on Robotics and Automation ({ICRA})}}.\hskip 1em plus 0.5em minus
  0.4em\relax IEEE, May 2015.

\bibitem{Meijering2004-lr}
E.~Meijering, M.~Jacob, J.-C.~F. Sarria, P.~Steiner, H.~Hirling, and M.~Unser,
  ``\BIBforeignlanguage{en}{Design and validation of a tool for neurite tracing
  and analysis in fluorescence microscopy images},''
  \emph{\BIBforeignlanguage{en}{Cytometry A}}, vol.~58, no.~2, pp. 167--176,
  Apr. 2004.

\bibitem{Rosenberg2007-qe}
A.~Rosenberg and J.~Hirschberg, ``{{V}-Measure}: A conditional {Entropy-Based}
  external cluster evaluation measure,'' in \emph{Proceedings of the 2007 Joint
  Conference on Empirical Methods in Natural Language Processing and
  Computational Natural Language Learning ({{EMNLP}-{C}o{NLL}})}.\hskip 1em
  plus 0.5em minus 0.4em\relax Prague, Czech Republic: Association for
  Computational Linguistics, June 2007, pp. 410--420.

\bibitem{Vinh2010-ti}
N.~X. Vinh, J.~Epps, and J.~Bailey, ``Information theoretic measures for
  clusterings comparison: Variants, properties, normalization and correction
  for chance,'' \emph{J. Mach. Learn. Res.}, vol.~11, no.~95, pp. 2837--2854,
  2010.

\bibitem{Steinley2004-le}
D.~Steinley, ``\BIBforeignlanguage{en}{Properties of the {Hubert-Arabie}
  adjusted rand index},'' \emph{\BIBforeignlanguage{en}{Psychol. Methods}},
  vol.~9, no.~3, pp. 386--396, Sept. 2004.

\bibitem{Fowlkes1983-an}
E.~B. Fowlkes and C.~L. Mallows, ``A method for comparing two hierarchical
  clusterings,'' \emph{J. Am. Stat. Assoc.}, vol.~78, no. 383, pp. 553--569,
  Sept. 1983.

\bibitem{Romano2016-mn}
S.~Romano, N.~X. Vinh, J.~Bailey, and K.~Verspoor, ``Adjusting for chance
  clustering comparison measures,'' \emph{J. Mach. Learn. Res.}, vol.~17,
  no.~1, pp. 4635--4666, Jan. 2016.

\bibitem{Luperto2019-pb}
M.~Luperto and F.~Amigoni, ``Extracting structure of buildings using layout
  reconstruction,'' in \emph{Intelligent Autonomous Systems 15}.\hskip 1em plus
  0.5em minus 0.4em\relax Springer International Publishing, 2019, pp.
  652--667.

\end{thebibliography}

\end{document}